\documentclass{sigkddExp}
\usepackage{graphicx}
\usepackage{subcaption}
\usepackage{soul}

\newcommand{\La}{\mathcal{L}}

\newcommand{\Ia}{\mathcal{I}}

\newcommand{\se}{u}
\newcommand{\te}{v}
\newcommand{\ti}{t}
\newcommand{\ac}{a}

\begin{document}
%
% --- Author Metadata here ---
% -- Can be completely blank or contain 'commented' information like this...
%\conferenceinfo{WOODSTOCK}{'97 El Paso, Texas USA} % If you happen to know the conference location etc.
%\CopyrightYear{2001} % Allows a non-default  copyright year  to be 'entered' - IF NEED BE.
%\crdata{0-12345-67-8/90/01}  % Allows non-default copyright data to be 'entered' - IF NEED BE.
% --- End of author Metadata ---

\title{TIES: Temporal Interaction Embeddings For Enhancing Social Media Integrity At Facebook}
\numberofauthors{3}
\author{
\alignauthor Nima Noorshams \\
	\affaddr{Core Data Science}\\
       \email{nshams@fb.com} 
\alignauthor Saurabh Verma\\
	\affaddr{Core Data Science}\\
       \email{saurabh08@fb.com}
\alignauthor Aude Hofleitner\\
	\affaddr{Core Data Science}\\
       \email{aude@fb.com}
}

\date{1 November 2019}
\maketitle

\begin{abstract}

Since its inception, Facebook has become an integral part of the online social community. People rely on Facebook to make connections with others and build communities. As a result, it is paramount to protect the integrity of such a rapidly growing network in a fast and scalable manner. In this paper, we present our efforts to protect various social media entities at Facebook from people who try to abuse our platform. We present a novel Temporal Interaction EmbeddingS (TIES) model that is designed to capture rogue social interactions and flag them for further suitable actions. TIES is a supervised, deep learning, production ready model at Facebook-scale networks. Prior works on integrity problems are mostly focused on capturing either only static or certain dynamic features of social entities. In contrast, TIES  can capture both these variant behaviors in a unified model owing to the recent strides made in the domains of graph embedding and deep sequential pattern learning. To show the real-world impact of TIES, we present a few applications especially for preventing spread of misinformation, fake account detection, and reducing ads payment risks in order to enhance Facebook platform's integrity.

\end{abstract}
\section{Introduction}~\label{sec:introduction}

%It’s been estimated that 5\% of Facebook's monthly active users between Q4-2018 and Q1-2019 belonged to fake accounts ~\cite{tr_fa}. These accounts pose a potential risk to the online community. They can be used to scam unsuspecting users~\cite{kurtzleben2018, mynamar_violance}.

People use online social media such as Facebook to connect with family and friends, build communities, and share experiences every day. But the rapid growth of social media in recent years has introduced several challenges. First is the rise of fake and inauthentic accounts, which pose potential threats to the safety of online communities. Second is the rise of threatening and/or disparaging content such as hate-speech, misinformation, bullying, terrorist propaganda, etc. These can occur both through authentic as well as inauthentic accounts. In Q1-2019 Facebook has acted on 4 million and 2.6 million pieces of content for violating hate-speech and bullying policies, respectively~\cite{tr_hate, tr_bully}. We broadly refer to these as \textit{social media integrity} challenges.

There is a volume of research on fake account detection in online social media. Graph propagation techniques to detect spams over user-following graphs have been proposed in ~\cite{yang2012analyzing, nilizadeh2017poised}. Integro~\cite{boshmaf2015integro} focuses on predicting “victims” using user-level features and performs a random walk style algorithm on a modified graph.  There have been other works that focus on utilizing the graph structure, by clustering and identifying cohorts of malicious actors that share common IP addresses and other common networking properties~\cite{stringhini2015evilcohort, zhaoetal2009, xiao2015detecting}.  Researchers in~\cite{li2016world} create graphs based on user activities in order to detect fake engagements. Similarly, hand-designed features based on activity has been used in~\cite{chen2018fakebuster}. 

There is also a volume of research on content-based integrity problems~\cite{schmidt2017, zubiaga2018}. Natural language processing techniques have been widely used for hate-speech and cyberbullying detection~\cite{chen2012, burnap2016, nobata2016, zhong2016, djuric2015}. Simple token and character-level n-grams are included in the feature set by~\cite{chen2012, burnap2016, nobata2016}. Word topic distribution using Latent Dirichlet Allocation has been used by~\cite{zhong2016} to detect cyberbullying on Instagram. Alternatively, paragraph embedding for hate-speech detection was proposed in~\cite{djuric2015}. Dinakar et al.~\cite{dinakar2012} presented a knowledge-based approach utilizing domain specific assertions in detecting hate-speech. In~\cite{hosseinmardi2015, zhong2016} authors combine image and other media features with text, essentially incorporating context through multimodal information. User meta-data such as the violation history, number of profane words in prior comments, gender, etc. have also been shown predictive~\cite{dadvar2013, waseem2016}. More recently, deep-learning has been used to fight child pornography~\cite{vitorio2018}, hate-speech~\cite{gamback2017, badjatiya2017}, and misinformation~\cite{ma2016, yu2017}.

\begin{figure*}[!t]
    \centering
    \begin{subfigure}[b]{0.4\textwidth}
       \hspace{3em}\includegraphics[width=0.8\textwidth]{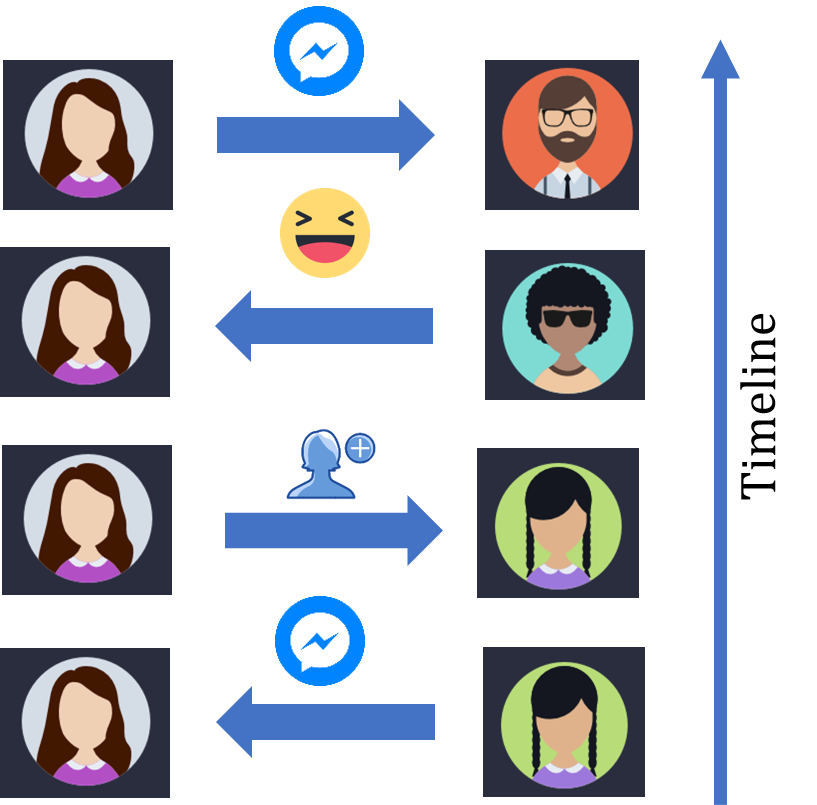}
        \caption{user-user interactions for fake account detection.}
        \label{fig:account_account_interactions}
    \end{subfigure}
    \quad
    \begin{subfigure}[b]{0.39\textwidth}
       \hspace{1em}\includegraphics[width=\textwidth]{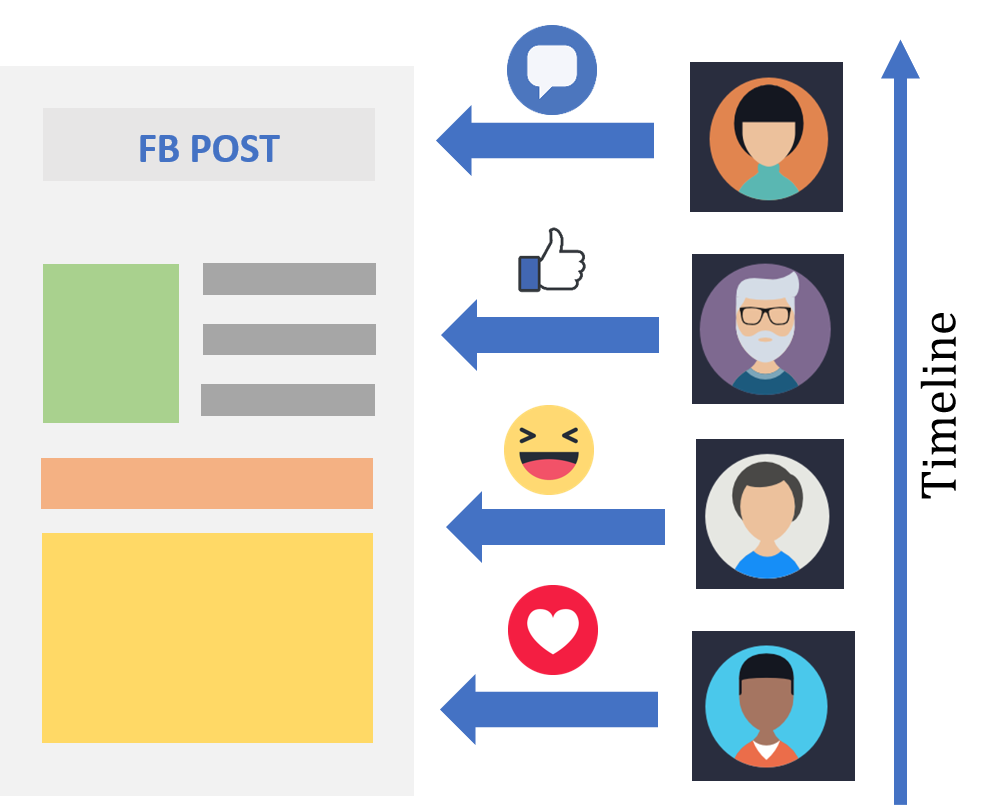}
        \caption{post-user interactions for hate-speech detection.}
        \label{fig:post_user_interactions}
    \end{subfigure}
\caption{Entities on social media interact with each other in numerous ways. Interactions generated by bad entities differ from normal entities. We can enhance the platform integrity by capturing/encoding these interactions.}
 \label{fig:interactions}
\end{figure*}

The majority of previous approaches tackling integrity challenges are static in nature. More specifically, they utilize engineered user-level, graph, or content features that do not alter in time. However, entities on social media (accounts, posts, stories, Groups, Pages, etc.) generate numerous interactions from other entities over time (see Figure~\ref{fig:interactions}). For instance, 
\begin{itemize}
\item posts get likes, shares, comments, etc. by users, or
\item accounts send or reject friend requests, send or block messages, etc. from other accounts.
\end{itemize}
These temporal sequences can potentially reveal a lot about entities to which they belong. The manner in which fake accounts behave is different from normal accounts. Hateful posts generate different type of engagements compared to regular posts. Not only the type but also the target of these engagements can be informative. For instance, an account with history of spreading hate or misinformation sharing or engaging positively with a post can be indicative of a piece of questionable content.

In this work, we present Temporal Interaction EmbeddingS (TIES), a supervised deep-learning model for encoding interactions between social media entities for integrity purposes. As its input, TIES takes a sequence of (source, target, action) in addition to miscellaneous source and target features. It then learns model parameters by minimizing a loss function over a labeled dataset. Finally, it outputs prediction scores as well as embedding vectors. There has also been other works on temporal interaction networks and embeddings. Recently and simultaneously to our work, JODIE, a novel embedding technique to learn joint user and item embeddings from sequential data, was proposed~\cite{kumar2019predicting}.  While these authors apply their algorithm to the problem of detecting banned accounts from Wikipedia and Reddit, which is similar to the problem of fake account detection, their work is different from ours. For example, they apply two recurrent neural networks to learn joint user-item embeddings based on temporal interaction sequences, while we just want to learn the embedding for the source entities. In order to leverage the target entities’ current state, we also use pre-existing embeddings or application specific feature sets which JODIE does not.  A notable limitation of JODIE and other existing approaches is scalability. On Facebook, we have billions of accounts and trillions of interactions per day. Therefore, scalability and reasonable computational costs are of utmost importance. 

% It is computationally prohibitive to learn or update each user/item embeddings at each timestep.

%JODIE also introduces a projection operator that estimates the embedding of a user at any time in the future in order to predict future user-item interactions.

%Despite the differences between our paper and JODIE, we provide an analysis of the performance of both algorithms in the experimental section of this paper. 

The remainder of the paper is organized as follows. We begin in Section~\ref{sec:problem_model} with the problem formulation and description of the model architecture. In Section~\ref{sec:applicatoins}, we discuss a couple of integrity case studies and results. Finally, we conclude the paper with discussions in Section~\ref{sec:discussions}.

%\clearpage

%\input{sections/related_work.tex}
\section{Protecting Facebook Social Media Integrity}\label{sec:problem_model}

We first start by providing a mathematical formulation for solving various social media integrity problems encountered on Facebook's platform and subsequently present the TIES model in detail. 

\subsection{Integrity Problem Formulation}\label{sec:problem}

At Facebook, we are interested in verifying the integrity of various social media entities such as accounts, posts, Pages, Groups, etc. As mentioned earlier, in this work we exploit the interaction information between such entities to determine their integrity. We refer to an entity under inspection as source (denoted by $\se$) while other interacted entities are referred as targets (denoted by $\te$). Suppose at time $\ti$, the source entity $\se$ interacts with target entity $\te_\ti$ by taking an action\footnote{Actions could originate from either source or target.} $\ac_\ti$ such as receiving a friend request, sending an event invite, or liking a post. We might also have source and target specific features $f_t$ at each timestamp (e.g. text or image related features or time gaps between consecutive actions). As a result, we will have a sequence of temporal interactions represented by $\Ia = \{(\se, \te_1, \ac_1, f_1), (\se, \te_2, \ac_2, f_2), \dots, (\se, \te_T, \ac_T, f_T)\}$. Based on the interaction sequence $\Ia$, TIES determines the integrity of $\se$, for instance, how likely is the entity to be a fake account or a hateful post. We do so by training a supervised model using a labeled training set $\{(\Ia_k, l_k)\}_{k=1}^{N}$, where $l_k$ is the ground truth label for the sequence $\Ia_k$. Thus, in our framework solving social media integrity is formulated as a  sequential or temporal learning problem. 

\subsection{TIES Model}\label{sec:model}

\begin{figure}[!t]
	\centering
	\includegraphics[width=1.0\linewidth]{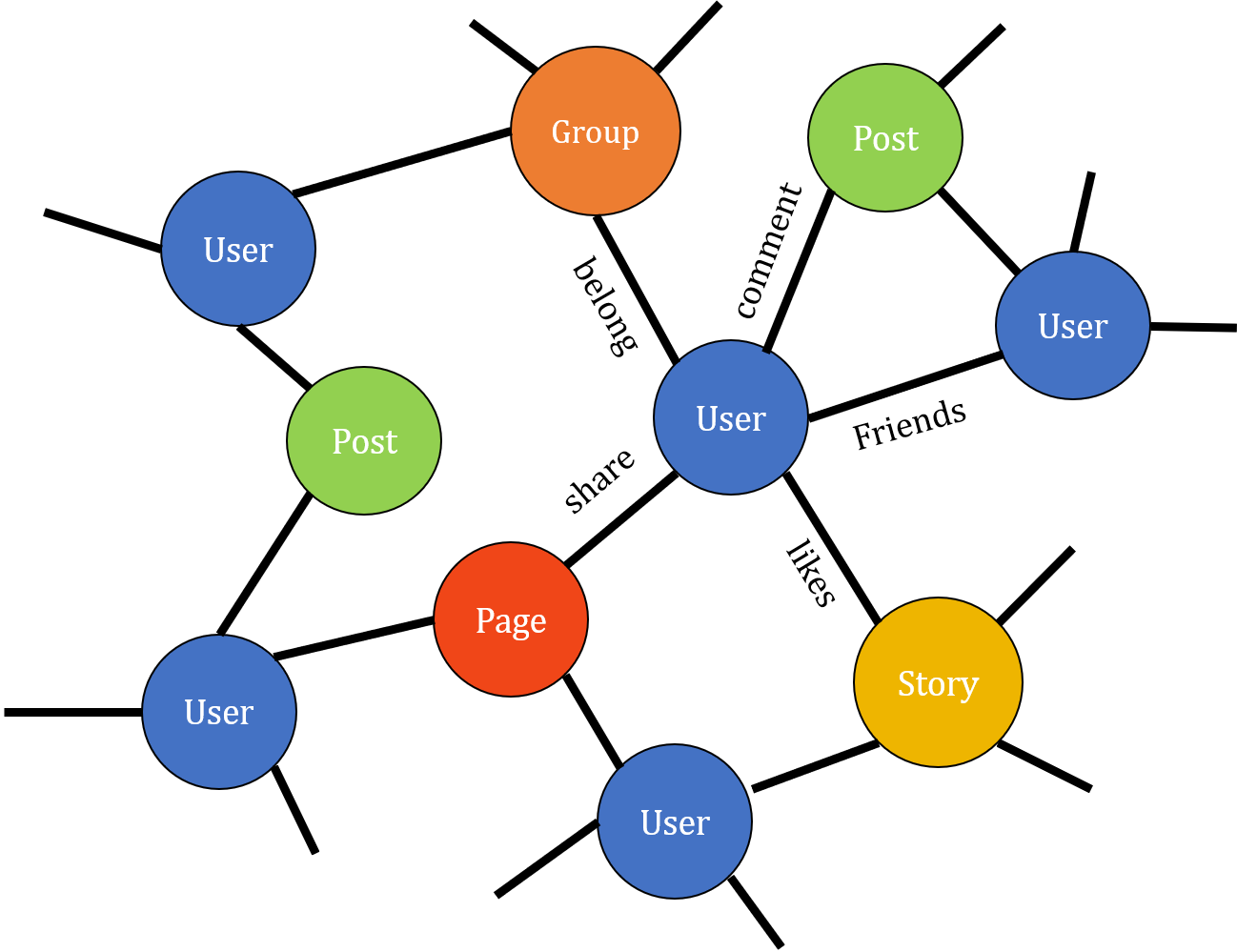}
	\caption{Facebook social media entity graph. To capture the static behavior of entities, Pytorch-BigGraph system is used to compute graph embeddings. These vectors are treated as pre-trained embeddings for the TIES model. }
	\label{fig:fb-graph}
\end{figure}

\begin{figure*}[t]
	\centering
	\includegraphics[width=0.95\linewidth]{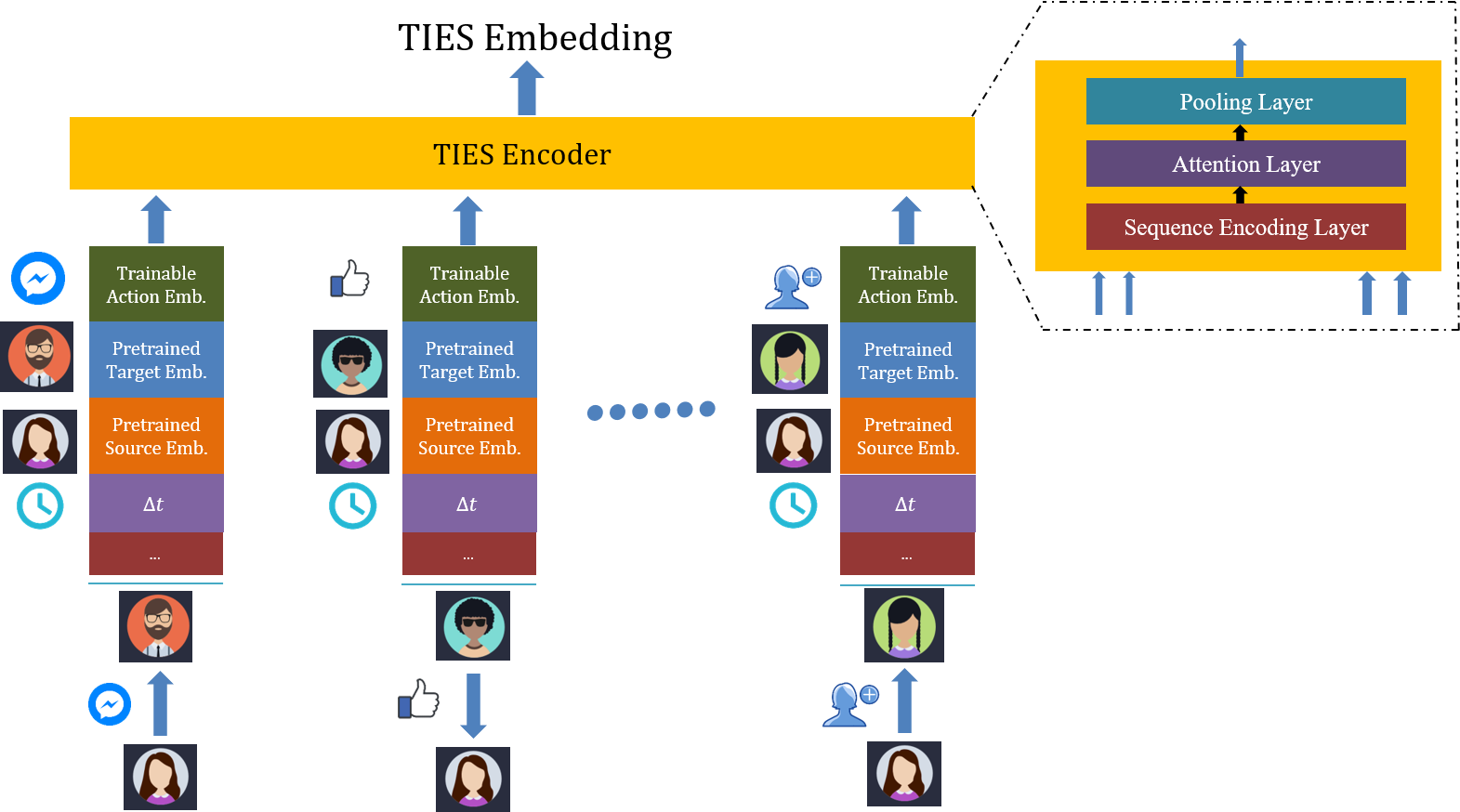}
	\caption{\textbf{TIES Model Architecture}: At each time step, the (source, target, action) triplet is converted into a feature vector that consists of trainable action embeddings, pre-trained source and target embeddings, and other miscellaneous features. The feature vectors are then fed into a deep sequential learning model to capture the dynamic behavior of entities.}
	\label{fig:ties-model1}
\end{figure*} 

At the core of the TIES model, there are two types of embeddings: 1) graph based and 2) temporal based. These embeddings are constantly trained to capture the ongoing behavior of the entities. We first make use of a large-scale Facebook graph to capture the static behavior of the entities. Subsequently, these graph-based embeddings are used to initialize the temporal model that captures the dynamic behavior of the entities. This is a distinguishing feature of the TIES that has not been explored before in prior integrity works. We now describe these parts in more detail.

% At the core of the TIES model, there are two  types of entity embeddings that are constantly learned for capturing ongoing intentions or behaviors 1) Graph Based Embeddings 2) Temporal Based Embeddings. The TIES model first makes use of a large scale Facebook entity graph to capture the static behavior in form of graph based embeddings. Secondly, temporal based embeddings are computed to capture the dynamic behavior of entities. We integrate the entity static behavioral information in the   model  by initializing it  with prior  learned graph  embeddings which results in a dramatic performance boost. This is a distinguishing feature of the TIES model that has not been explored before in prior integrity works. Both of these embeddings are discussed in detail below.

%\textcolor{red}{Describe here in short how graph based embedding and temporal based embeddings are integrated in TIES model before going into details}.

%
%\begin{itemize}
%	\item Discuss the  importance of  pretrained embeddings. What information do they capture?
%	One of the most insightful 
%	
%	\item Stress that this is missing from all previous work.
%	\item How we going to compute them in scalable fashion. Discuss using Pytorch Big Graph~\cite{lerer2019pytorch}.
%		\item Make a deep dive on explaining the actual model component.
%\end{itemize}

\subsubsection{Graph Based Embeddings}\label{sec:graph_embeddings}

One of the novel components of TIES model is making use of a large scale graph structure formed by various social media entities beside the dynamic interaction information. In social networks, entities may be connected to each other through friend relationships or belong to the same groups. Such prior knowledge can capture the `static' nature of various entities. It should be noted that even though these static relations (such as friend relationships, group membership, etc.) are not truly static, they vary at a much lower pace compared to other more `dynamic' interactions (such as post reactions, commenting, etc.). Past studies mainly focus either on static or dynamic behavior but do not account for both in the model. Moreover, the scale of the graph structure considered in this work is much greater than in previous works and thus presents unique challenges. 

Let $G = (V, R, E)$ denote a large-scale multi-relations graph formed by social media entities (see Figure~\ref{fig:fb-graph}). Here, $V$ denotes a set of nodes (i.e., entities), $R$ is a set of relations and $E$ denotes a set of edges. Each edge $e = (s, r, d)$ consists of a source $s$, a relation $r$, and a destination $d$, where $s, d\in V$ and $r\in R$. In order to learn graph embeddings for entities in the graph $G$, we utilize the PyTorch-BigGraph (PBG)~\cite{lerer2019pytorch} distributed system due to its scalability to billions of nodes and trillions of edges. This is essential for solving our real-world needs. 

Suppose $\theta_s, \theta_r, \theta_d$ are trainable parameter vectors (embeddings) associated with source, relation, and destination. PBG assigns a score function $f(\theta_s, \theta_r, \theta_d)$ to each triplet, where higher values are expected for $(s, r, d) \in E$ as opposed to $(s, r, d) \notin E$. PBG optimizes a margin-based ranking loss (shown below) for each edge $e$ in the training data. A set of judiciously constructed negative edges $e'$ are obtained by corrupting $e$ with either a sampled source or destination node
\begin{align*}
\ell = \sum_{e \in G}\sum_{e' \in S_{e}^{'}} \max\{ f(e) -f(e') +\lambda, 0 \}
\end{align*}
Here $\lambda$ is the margin hyperparameter and $S_{e}^{'}= \{(s', r, d) | s'\in V\} \cup \{(s, r, d') | d'\in V\} $. Finally, entity embeddings and relation parameters are learned by performing mini-batch stochastic gradient descent. More details about these embeddings can be found in~\cite{lerer2019pytorch}. The PBG-trained embeddings on a large-scale graph with billions of nodes and trillions of edges are fed as pre-trained source and target embeddings to the temporal model, to which we now turn.

%We smartly  exploit $G$ to avail the benefits of  structural information  in  the form of graph embeddings which are responsible for capturing the  static behavior of social media entity.  
%For an instance in Facebook social network, nodes could be a group, page or post and edges could represent a friend-relationship or connection  made by commenting on the same page or post. 

%\textcolor{red}{Should we move below paragraph to application section?}

%To that end, the Facebook graph can be seen as a very large scale, multi-relations graph with over 2 billion nodes as user type entities and millions of other entity types as group, pages or posts. Among these entities trillion of multi-relation edges are formed due to   different types of connections such as friendship or belonging to the same group. Using PBG system, graph embeddings (with $d = 100$ dimension) are learned for each entity in Facebook graph and treat them as pre-trained embeddings for downstream applications.  As a result, we have pretrained source and target embeddings representing source media entities and ready to be consumed as the  part of our TIES model input.  

\subsubsection{Temporal Based Embeddings} \label{sec:temporal_embeddings}

Temporal based embeddings are designed to capture the information encoded in the sequence of interactions $\Ia$ as discussed in Section~\ref{sec:problem}. Consider  all the interactions of a source entity $u$ in a given window of time period. Suppose at  $t$ time, source entity $u$  interacted with target entity $v_t$ by performing an action $a_t$.  This whole interaction at time $t$ is encoded into a single feature vector  as follows (see Figure~\ref{fig:ties-model1}): \\

\begin{enumerate}
\item \textbf{Action Features}: Action $a$, for instance commenting or liking, is represented by a fixed size vector  that will be learned during the training process and initialized as random (also referred as trainable embedding). Depending upon the task, the same type of action can learn different embeddings and can have multiple meanings based on the context. As mentioned earlier, one can also encode the direction information in action by splitting it into two directional events as  $a-$send or $a-$receive.

\item \textbf{Source Entity Features}: Source entity $u$ is represented by a pre-trained embedding. More specifically, we utilize graph-based embeddings obtained from the PBG system as described in Section~\ref{sec:graph_embeddings}.  One can further finetune the pre-trained embeddings in the TIES model, but this is only possible if the number of unique source entities is not greater than a few million (due to computational cost).

\item \textbf{Target Entity Features}: Similar to the source entity, the target entity $v$ is also represented by a pre-trained embedding (if available) or by a trainable embedding (if problem dimension allows i.e., limited to few millions). Multiple types of pre-trained target embeddings can be utilized in the same TIES model.

\item \textbf{Miscellaneous features}:  We can also encode useful time-related information such as rate of interaction via $\Delta_t = t_{i+1} - t_i$  (may need to normalize the range appropriately). Rate of interaction is an important signal for detecting abusiveness. Other features like text or images can also be plugged into TIES in similar manner.
	
\end{enumerate}

All these features are packaged into a single feature vector by performing an aggregation operation. We obtain a single embedding  capturing the full interaction information in
\begin{align*}
\mathbf{x}_t = \mathbf{e}(u)\odot \mathbf{e}(v_t) \odot \cdots  \odot \Delta_t, 
\end{align*}
where $\odot$ is a aggregation operator, $\mathbf{e}(\cdot) $ represents the entity embedding and $\mathbf{x}_t$ is the resultant embedding obatined at time $t$.  In our case, we simply choose concatenation as the aggregation operation. Next, we pass the sequence of these temporal embeddings i.e.,  $\mathbf{X}=[\mathbf{x}_1, \mathbf{x}_2, ..., \mathbf{x}_T] \in \mathbb{R}^{T \times d}$ where $T$ is sequence length and		 $d$ is the input dimension, into a sequence encoder to yield  TIES embedding $\mathbf{z} \in \mathbb{R}^{h}$ as follows
\begin{align*}
\mathbf{z}= TIESEncoder(\mathbf{X}).\\
\end{align*}

\noindent \textbf{TIES Encoder}: It yields our final TIES embedding by capturing the temporal aspect present in the sequence of interaction embeddings. Our general purpose sequence encoder has the following components:

\begin{enumerate}
\item \textbf{Sequence Encoding Layer}:  This encoding layer transforms the input into a hidden state that is now aware of the interaction context in the sequence
\begin{align*}
\mathbf{H} = SeqEnocder(\mathbf{X}).
\end{align*}
Here, $\mathbf{H} \in \mathbb{R}^{T \times h}$ is the hidden-state matrix with $h$ as the dimension. We consider three types of sequence encoding layer in our TIES framework with varying training and inference costs and benefits:
	\begin{enumerate}
	\item \textbf{Recurrent neural networks}:  RNNs such as long short term memory networks (LSTM) 	are quite capable of capturing dependencies in a sequence~\cite{lipton2015rnnreview}. But they 	are inherently slow to train.
	\item \textbf{Convolutional neural networks}: 1D sequence CNNs can also capture sequential 		information but are limited to local context and need to have higher depth for capturing global context, depending on the task in hand~\cite{khan2019cnnreview}.
	\item \textbf{DeepSet}: When inputs are treated as sets and their order does not matter, we can use Deepsets as sequence encoders~\cite{zaheer2017deepset}. Here, we first pass each input in a sequence through an MLP (small neural network) and then perform a sum operation followed by another MLP layer, yielding a single embedding layer.
	\end{enumerate}
Besides the application in-hand and deployment challenges, the choice among RNN, CNN and DeepSet depends on the tradeoff between performance and inference time. Due to its recurrent nature, RNN is expensive while DeepSet has the lowest inference time in production.

\item \textbf{Attention Layer}: Attention Layer~\cite{vaswani2017attention} can be used to weigh the embeddings differently according to their contribution towards specific tasks. Attention values can also be used to visualize which part of the interaction sequence is being focused more than the others and that can provide more interpretable outcome. The output is given by
\begin{align*}
\mathbf{Z} =Attention(\mathbf{H}), 
\end{align*}
where $\mathbf{Z} \in \mathbb{R}^{T \times h}$ is the attention layer output.

\item \textbf{Pooling Layer}: A final pooling layer such as mean, max, or sum operation is used to yield a single embedding for the whole interaction sequence. Here, we have 
$$\mathbf{z} =Pooling(\mathbf{Z})$$ 
where $\mathbf{z} \in \mathbb{R}^{h}$ is the output of the pooling layer and serves as the final TIES embeddings.
\end{enumerate}

\subsubsection{Loss Function} Parameters of TIES are learned based on the task labels associated with each training sequence $\Ia_k$, $k=1, 2, ..., N$. For instance, in case of abusive account detection we have a training set of sequences labeled as either abusive or benign and binary cross-entropy can be considered as the loss function. In general, TIES embedding $\mathbf{z}$ is fed to  the feed-forward neural network for learning the parameters in end-to-end fashion  as follows,
$$ \ell = \sum_{i=1}^{N} \La(\mathbf{X}_i, f(\mathbf{z}_i)) $$
where $\ell$ is the overall task loss, $\La$ is the loss function, $f(\cdot)$ is the feed-forward neural network, $\mathbf{X}_i \in  \mathbb{R}^{T \times d}$ is the $i^{th}$ input training sequence, $\mathbf{z}_i \in  \mathbb{R}^{h}$ is the corresponding learned TIES embedding and $N$ is the total number of training samples. Depending upon the task, different metrics are taken into consideration and class-imbalance issue is handled by weighting the classes properly. This completes the overall description of the TIES model architecture.

\section{Facebook Integrity Applications and Results}~\label{sec:applicatoins}

In this section, we briefly describe our implementation of TIES, introduce some of its integrity applications, and use cases. These applications cover a wide range of issues from content-based to account-based. 

\subsection{Production Implementation}

Our framework is built on PyTorch, more specifically TorchScript to streamline productionization~\cite{jit_doc}. Training is generally performed on GPUs and when needed we use DistributedDataParallel provided by PyTorch for multi-GPU training. Inference on the other hand is performed in parallel on up to 200 machines (CPUs suffice during inference). To maintain relative consistency in embedding vectors overtime, we use warm-start---that is, initializing the model with a previously trained one. For our experiments, we use Adam optimizer with learning rate 0.0005 and clip gradients at 1. To mitigate overfitting we use dropout with probability 0.1. Finally, we weight positive samples such that the datasets are balanced.

\subsection{Preventing Spread of Misinformation}\label{sec:misinfo}

False news can spread rapidly, either intentionally or unintentionally, by various actors. Therefore, proactive identification of misinformation posts is a particularly important issue. It is also very challenging as some of such posts contain material that are not demonstrably false but rather are designed to be misleading and/or reflecting only one side of a story. 

Some of the existing techniques train classifiers on carefully crafted features~\cite{hamidian2016}. There are also approaches that detects ``rumors" based on users' reactions to microblogs overtime~\cite{ma2016, yu2017}. To the best of our knowledge, users' histories (captured via graph-embeddings described in~\ref{sec:model}) and their interactions with posts have not been used to detect misinformation systematically. More recently, researchers at Facebook have devised a multimodal technique for identifying misinformation, inspired by the work of Kiela et al. ~\cite{kiela2018}. In the multimodal model, separate encoders are learned for various content modalities such as image, post-text, image-embedded-text, comments, etc. The encoded features are then collected into a set and pooled in an order-invariant way. Finally, the pooled vector is passed through a fully connected layer to generate predictions.
%\begin{figure}
%    \centering
%    \begin{subfigure}[b]{0.3\textwidth}
%        \includegraphics[width=\textwidth]{figs/result_misinfo-roc-auc-box}
%        \caption{}
%        \label{fig:misinfo_results_roc}
%    \end{subfigure}
%    \quad
%    %add desired spacing between images, e. g. ~, \quad, \qquad, \hfill etc. 
%     %(or a blank line to force the subfigure onto a new line)
%    \begin{subfigure}[b]{0.3\textwidth}
%        \includegraphics[width=\textwidth]{figs/result_misinfo-pr-auc-box}
%        \caption{}
%        \label{fig:misinfo_results_pr}
%    \end{subfigure}
%\caption{Model performances on the test dataset. Area under the curve box-plot for (a) ROC and (b) precision-recall. Combining TIES with content improves the performance, significantly.}
% \label{fig:misinfo_results_auc}
%\end{figure}

\begin{figure} 
	\centering
	\includegraphics[width=0.4\textwidth]{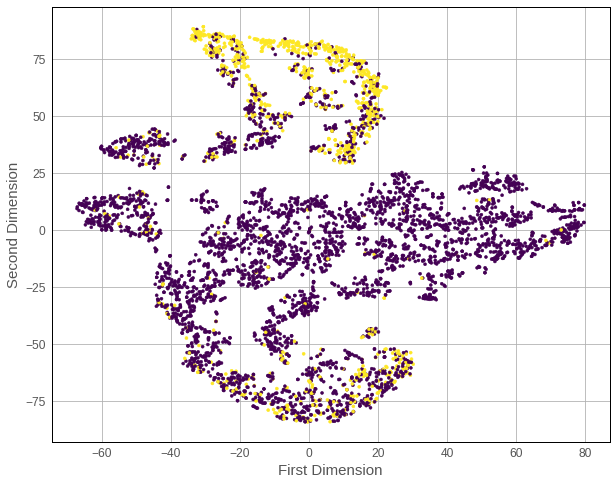}
        \label{fig:misinfo_results_embd_prj}
\caption{ 2-dimensional TSNE projection of the TIES embeddings for \textbf{misinformation}. Clearly, the misinformation posts (yellow) have a different distribution than the regular posts (purple).}
 \label{fig:misinfo_results_prj}
\end{figure}

%\begin{table}[h!]
%\begin{center}
%\begin{tabular}{|c|c|c|}
%\hline
 %Model & PR-AUC Median $\pm$ MAD \\ 
 %\hline
 %TIES-CNN & 0.6358 $\pm$ 0.0058 \\  
 %\hline
 %TIES-RNN & 0.6501 $\pm$ 0.0091 \\
 %\hline
  %TIES-Deepset & 0.6336 $\pm$ 0.0072 \\
 %\hline
 %Content-Only & 0.7488 $\pm$ 0.0052 \\
 %\hline
  %Content+TIES-CNN & 0.7874 $\pm$ 0.0035 \\
 %\hline
  %Content+TIES-RNN & 0.7962 $\pm$ 0.0050\\
 %\hline
 %Content+TIES-Deepset & 0.7924 $\pm$  0.0035\\
 %\hline
%\end{tabular}
%\end{center}
%\caption{Precision-Recall area under the curve median and median absolute deviation on the test dataset for \textbf{misinformation} detection. Combining TIES with content improves the performance significantly.}
%\label{table:misinfo_results_prauc}
%\end{table}

\begin{table}[h!]
\begin{center}
\begin{tabular}{|c|c|c|}
\hline
 Model & PR-AUC Median Gap $\pm$ MAD \\ 
 \hline
 TIES-CNN & -0.1130 $\pm$ 0.0110 \\  
 \hline
 TIES-RNN & -0.0987 $\pm$ 0.0143 \\
 \hline
  TIES-Deepset & -0.1152 $\pm$ 0.0124 \\
 \hline
  Content+TIES-CNN & 0.0386 $\pm$ 0.0087 \\
 \hline
  Content+TIES-RNN & 0.0474 $\pm$ 0.0102\\
 \hline
 Content+TIES-Deepset & 0.0436 $\pm$  0.0087\\
 \hline
\end{tabular}
\end{center}
\caption{Median Precision-Recall area under the curve difference with respect to the content-only model and median absolute deviation on the test dataset for \textbf{misinformation} detection. Combining TIES with content improves the performance significantly.}
\label{table:misinfo_results_prauc}
\end{table}

In the context of TIES, we have posts (sources) interacting with users (targets). Here, we consider a small set of interactions: like, love, sad, wow, anger, haha, comment, and share. Moreover, we use embeddings described in Section~\ref{sec:model} for source and target entities. For this experiment, we split our training dataset, consisting of 130K posts (roughly 10\% of which are labeled as misinformation), into \textit{train-1}, \textit{train-2}, and \textit{test} sets\footnote{We use two disjoint training sets to prevent overfitting.}. It should be noted that this dataset is sampled differently for positive and negative cases and does not reflect the accurate distribution of the posts on our platform. We then use the set \textit{train-1} to train a few TIES models (CNN, RNN, and Deepset). For all models, we set the interaction embeddings as well as hidden dimensions to 64. We consider sequences of length 512, where longer sequences are cropped from the beginning and shorter sequences are padded accordingly.  The CNN model consists of 2 convolution layers of width 5 and stride 1. The RNN model consists of a 1-layer bidirectional LSTM. And finally, the Deepset model consists of pre and post aggregation MLPs with one hidden layer of size 64. In addition to TIES, we train a multimodal model using post images and texts. Finally, we use the \textit{train-2} dataset to train a hybrid model, a simple logistic-regression with two features, TIES-score and multimodal-score. In order to specify confidence intervals, we run the aforementioned experiment on several train/test data split. Table~\ref{table:misinfo_results_prauc} illustrates the difference/delta performance for various models with respect to the content-only model on the \textit{test} dataset.

At this point a few observations are worth highlighting: First, TIES-RNN seems to be the best performing TIES model. Second, Deepset appears to be the least performing TIES model, perhaps highlighting the importance of the ordered sequences (as opposed to sets) in identifying questionable posts. Third, the content-only model seems to outperform interaction-only models. Finally, combining the interactions signal with content (hybrid models) improves the performance significantly.
% \begin{figure}
%   \centering
%    \begin{subfigure}[b]{0.22\textwidth}
%        \includegraphics[width=\textwidth]{figs/result_misinfo_action_prj}
 %       \caption{}
%        \label{fig:misinfo_results_action_prj}
 %   \end{subfigure}
 %   \quad
%    \begin{subfigure}[b]{0.22\textwidth}
 %       \includegraphics[width=\textwidth]{figs/result_misinfo_embd_prj}
 %       \caption{}
 %       \label{fig:misinfo_results_embd_prj}
 %   \end{subfigure}
%\caption{ 2-dimensional TSNE projection of the TIES embeddings for (a) actions, and (b) posts. Clearly, the misinformation posts (yellow) have a different distribution than the regular posts (purple).}
% \label{fig:misinfo_results_prj}
%\end{figure}
\begin{figure}
    \centering
    \begin{subfigure}[b]{0.18\textwidth}
        \includegraphics[width=\textwidth]{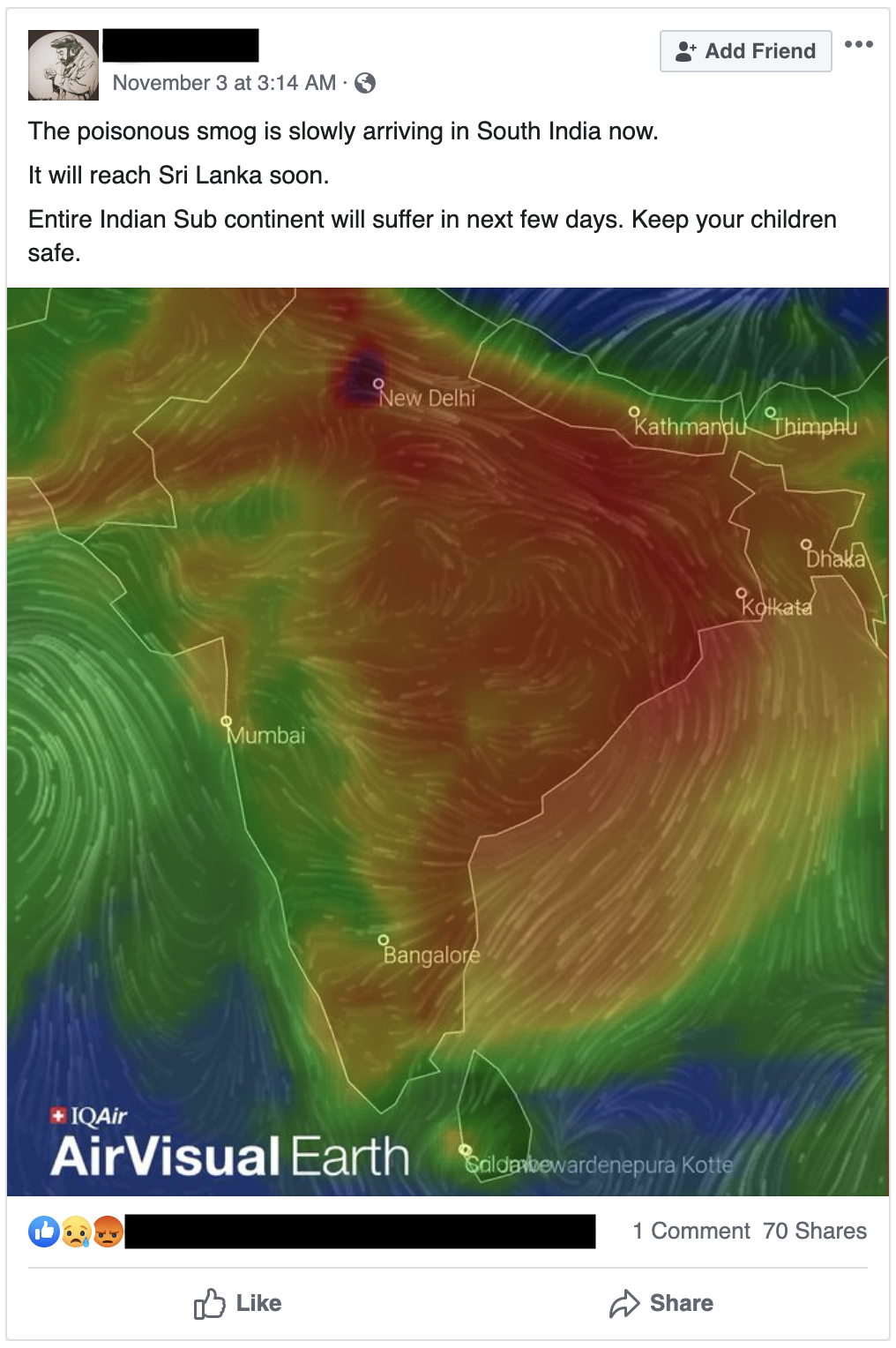}
        \caption{}
        \label{fig:misinfo_results_ex2}
    \end{subfigure}
    \quad
    \begin{subfigure}[b]{0.18\textwidth}
        \includegraphics[width=\textwidth]{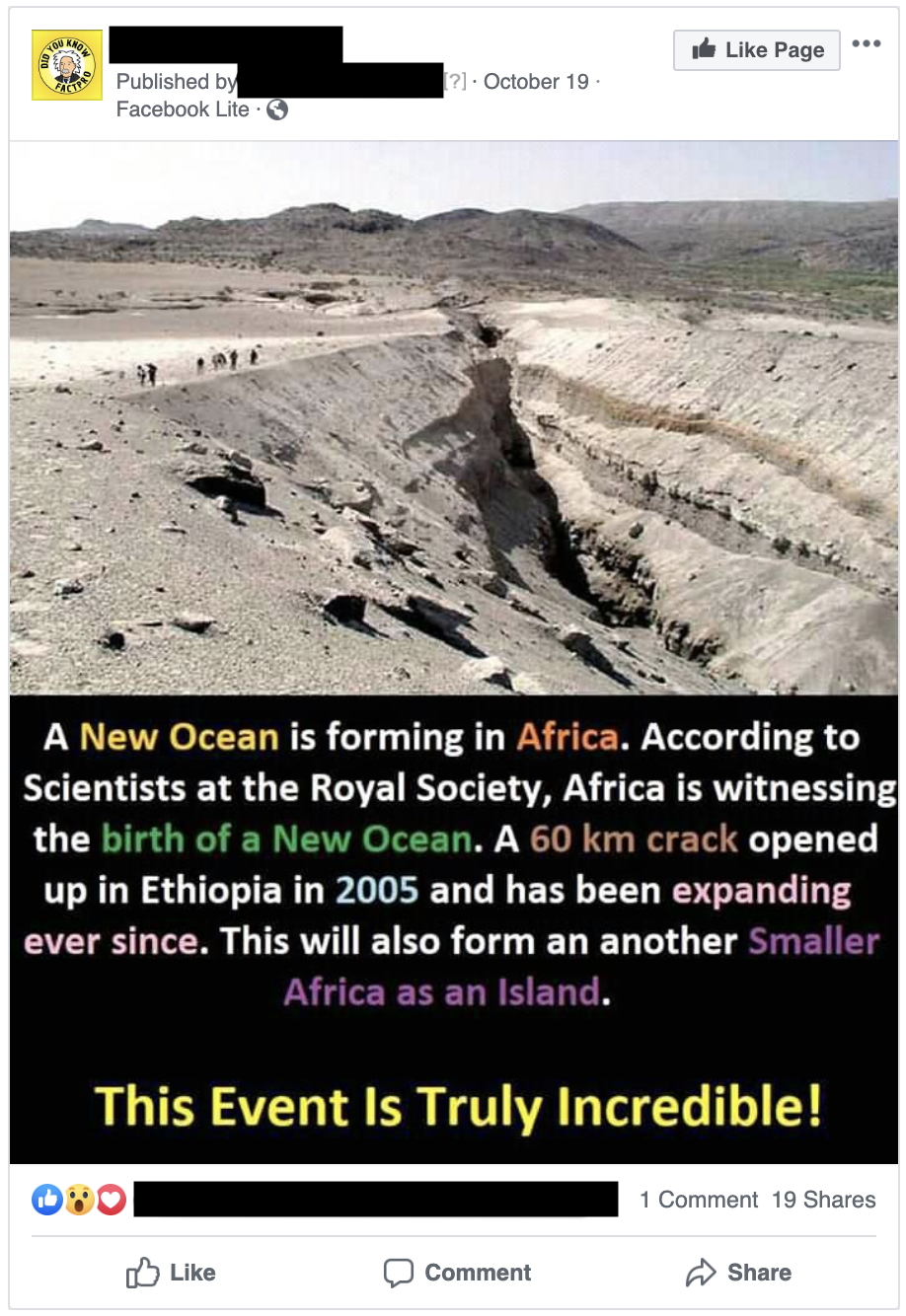}
        \caption{}
        \label{fig:misinfo_results_ex3}
    \end{subfigure}
\caption{Sample misinformation posts, as identified by independent human reviewers, with low content-only but high hybrid score.}
 \label{fig:misinfo_results_examples}
\end{figure}

Figure~\ref{fig:misinfo_results_prj} illustrates the TSNE projection of the post embeddings. It should be noted that the final hidden state of the TIES model is considered the source-entity and in this case post embeddings. Interestingly, the distribution of the misinformation posts (yellow dots), in the latent space, is clearly different from the regular posts (purple dots). Figure~\ref{fig:misinfo_results_examples} illustrates a couple of positive posts that were not identified by the content-only model but registered high scores in the hybrid (content+TIES) model.

%%%%%%%%%%%%%%%%%%%%%%%%%%%%%%%%%%%%%%%%%%%%%%%%%%%%
%%%%%%%%%%%%%%%%%%%%%%%%%%%%%%%%%%%%%%%%%%%%%%%%%%%%
\subsection{Detecting Fake Accounts and Engagements}\label{sec:fake_eng}

It is understood that fake-accounts are a potential security and integrity risk to the online community. Therefore, identifying and removing them proactively is of utmost importance.  Most existing approaches in detecting fake accounts revolve around graph propagation~\cite{yang2012analyzing, nilizadeh2017poised} or clustering of malicious actors and/or activities~\cite{stringhini2015evilcohort, zhaoetal2009, xiao2015detecting, li2016world}. Fakebuster~\cite{chen2018fakebuster} trained a model on activity related features. To the best of our knowledge, sequence of accounts and associated activities have not been used in fake account detection.

% As mentioned previously in Section~\ref{sec:introduction}, fake accounts are an enormous integrity challenge in social media. Roughly 5\% of Facebook's monthly active accounts are fake~\cite{tr_fa}. Even more accounts engage in inauthentic activities such as selling likes, follow, or spam users. Taking down these fake accounts and engagements is essential in safety and security of the online community. 

One of the fake engagement baseline models at Facebook is a multilayer classifier trained on over one thousand carefully engineered features. These features cover a range of signals from metadata, activity statistics, among others. In this experiment, we train a few TIES models and combine them with the output of the baseline classifier in order to evaluate the effectiveness of the sequence data. Here, sources are accounts engaging with various targets. Targets on the other hand could be other accounts, posts, or pages. As engagements, we consider a set of 44 sentry-level actions that includes liking a post, following a page, messaging a user, friending, etc. As source and target features, we use graph-based embeddings described in Section~\ref{sec:model} for accounts, as well as posts and pages creators, where applicable. 

Our dataset consists of 2.5M accounts with 80/20 good/bad (fake) split. It should be emphasized that this dataset is sampled differently for positive and negative cases and does not reflect the accurate distribution of fake accounts on our platform. We randomly divide this dataset into \textit{train-1}, \textit{train-2}, and \textit{test} sets consisting of 2M, 250K, and 250K accounts, respectively. We then use the set \textit{train-1} to train a few TIES models similar to the previous experiment described in Section~\ref{sec:misinfo}. The main difference is that here the CNN model has 3 convolution layers of width 5 and stride 1, and the RNN model consists of a 2-layer bidirectional LSTM. Subsequently, we use the \textit{train-2} dataset to train a logistic-regression with two features, TIES-score and baseline-score. This experiment is repeated on several train/test splits in order to calculate confidence intervals. The performance gap between various models and the baseline on the \textit{test} dataset are illustrated in Table~\ref{table:fe_results_prauc}. 
%\begin{figure}
%    \centering
%    \begin{subfigure}[b]{0.30\textwidth}
%        \includegraphics[width=\textwidth]{figs/result_fe-roc-auc-box}
%        \caption{}
%        \label{fig:fe_results_roc}
%    \end{subfigure}
%    \quad
%    %add desired spacing between images, e. g. ~, \quad, \qquad, \hfill etc. 
%     %(or a blank line to force the subfigure onto a new line)
%    \begin{subfigure}[b]{0.30\textwidth}
%        \includegraphics[width=\textwidth]{figs/result_fe-pr-auc-box}
%        \caption{}
%        \label{fig:fe_results_pr}
%    \end{subfigure}
%\caption{Model performances on the test dataset. Area under the curve box-plot for (a) ROC and (b) precision-recall. Combining TIES with prod improves the performance.}
% \label{fig:fe_results_auc}
%\end{figure}

%\begin{table}[h!]
%\begin{center}
%\begin{tabular}{|c|c|c|}
%\hline
 %Model & PR-AUC Median $\pm$ MAD \\ 
 %\hline
 %TIES-CNN & 0.7424 $\pm$ 0.0016 \\  
 %\hline
 %TIES-RNN & 0.7469 $\pm$ 0.0005 \\
 %\hline
 % TIES-Deepset & 0.7133 $\pm$ 0.0011 \\
 %\hline
 %Prod & 0.7990 $\pm$ 0.0006 \\
 %\hline
 % Prod+TIES-CNN & 0.8080 $\pm$ 0.0008 \\
 %\hline
 % Prod+TIES-RNN & 0.8100 $\pm$ 0.0008 \\
 %\hline
 %Prod+TIES-Deepset & 0.8045 $\pm$ 0.0006 \\
 %\hline
%\end{tabular}
%\end{center}
%\caption{Precision-Recall area under the curve median and median absolute deviation on the test dataset for \textbf{fake engagement} detection. Combining TIES with prod improves the performance.}
%\label{table:fe_results_prauc}
%\end{table}

\begin{table}[h!]
\begin{center}
\begin{tabular}{|c|c|c|}
\hline
 Model & PR-AUC Median Gap $\pm$ MAD \\ 
 \hline
 TIES-CNN & -0.0566 $\pm$ 0.0022 \\  
 \hline
 TIES-RNN & -0.0521 $\pm$ 0.0011 \\
 \hline
  TIES-Deepset & -0.0857 $\pm$ 0.0017 \\
 \hline
  Baseline+TIES-CNN & 0.0090 $\pm$ 0.0014 \\
 \hline
  Baseline+TIES-RNN & 0.0110 $\pm$ 0.0014 \\
 \hline
 Baseline+TIES-Deepset & 0.0055 $\pm$ 0.0012 \\
 \hline
\end{tabular}
\end{center}
\caption{Median Precision-Recall area under the curve difference with respect to the baseline model and median absolute deviation on the test dataset for \textbf{fake engagement} detection. Combining TIES with the baseline improves the performance.}
\label{table:fe_results_prauc}
\end{table}

\begin{figure} 
	\centering
	\includegraphics[width=0.4\textwidth]{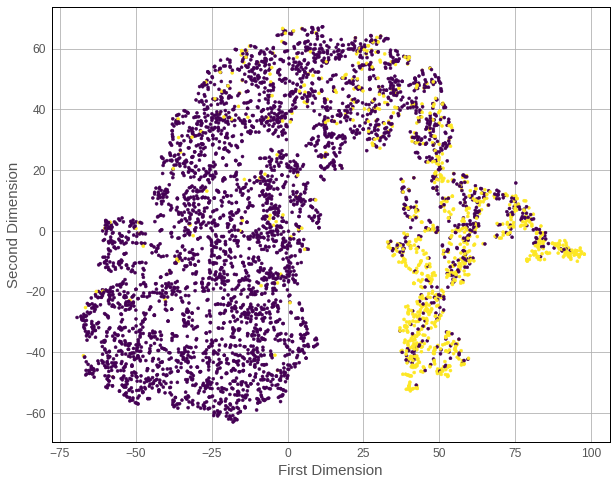}
        \label{fig:misinfo_results_embd_prj}
\caption{ 2-dimensional TSNE projection of the TIES embeddings for \textbf{fake engagements}. Clearly, the fake accounts (yellow) have a different distribution than the regular accounts (purple).}
 \label{fig:fe_results_prj}
\end{figure}

Our observations are largely consistent with the ones made in Section~\ref{sec:misinfo}. Namely, TIES-RNN appears to be the best performing TIES model, baseline outperforms TIES, and as an additional feature, TIES can provide a boost to the baseline model. The fact that baseline outperforms TIES is not surprising, as it includes a lot more information through over 1000 carefully engineered features. Moreover, gains from TIES appear to be small but they are statistically significant and outside the confidence interval. \textit{It should also be noted that, at the scale of Facebook, even a couple of percentage points improvement in recall for the same precision translates into significant number of additional fake accounts being caught.} Figure~\ref{fig:fe_results_prj} illustrates the 2-d projection of the TIES embeddings and as expected the distribution of the fake accounts (yellow) is quite different from nor mal accounts (purple).

%%%%%%%%%%%%%%%%%%%%%%%%%%%%%%%%%%%%%%%%%%%%%%%%%%%%
%%%%%%%%%%%%%%%%%%%%%%%%%%%%%%%%%%%%%%%%%%%%%%%%%%%%
\subsection{Reducing Ads Payment Risks}\label{sec:payment}

% Benefits of such an advertising platform for users is also evident in excellent but free services, relevant as opposed to random ads, and more importantly, millions of additional jobs created all over the world. 

With more than two billion monthly active users, Facebook enables small- and medium-sized  businesses to connect with a large audience. It allows them to reach out to their target audiences in an efficient way and grow their businesses as a result~\cite{fb_smb}. 

\begin{figure} 
	\centering
	\includegraphics[width=0.4\textwidth]{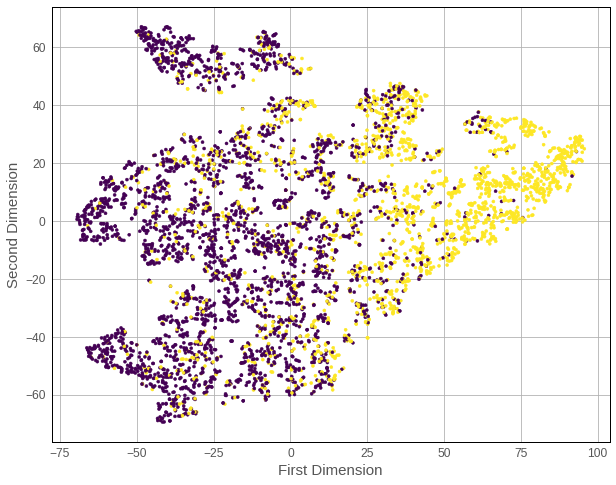}
        \label{fig:payment_results_embd_prj}
\caption{ 2-dimensional TSNE projection of the TIES embeddings for \textbf{ads payment risk}. Clearly, accounts with unpaid fees (yellow) have a different distribution than the regular accounts (purple).}
 \label{fig:payment_results_prj}
\end{figure}

Some of the integrity issues facing this advertising platform include fraudulent requests and unpaid service fees or substantial reversals. Researchers at Facebook have devised various models to prevent such abuses. These models generally include thousands of carefully crafted features that cover a wide range of information sources, such as user metadata, activity history, etc. In order to test TIES' viability for identifying bad accounts that have failed to pay fees, we train a few models using interaction signals that generally include details about the account and associated payment trends. Here, sources are ads accounts, source features are graph-based embeddings, and targets are generally nulls. We follow the settings in the previous two experiments: we split our dataset into \textit{train-1}, \textit{train-2}, and \textit{test} sets consisting of roughly 200K, 10K, and 10K accounts, respectively. The datasets are sampled such that we have roughly the same number of bad and good accounts. We then train TIES-CNN (2 layers of width 5), TIES-RNN (1 layer biLSTM),  and TIES-Deepset (pre and post aggregation MLPs with 1 hidden-layer of size 64) as well as the baseline model on the set \textit{train-1}. We then use \textit{train-2} dataset to combine baseline and TIES scores via a simple logistic regression and finally test the outcome on the \textit{test} dataset. In order to calculate the confidence intervals, we repeat this experiment on several random train/test splits. Precision-recall area under the curve gaps with respect to the baseline model are illustrated in Table~\ref{table:payment_results_prauc}. Figure~\ref{fig:payment_results_prj}, on the other hand, demonstrates the 2-d projection of embedding vectors and as expected the distribution of bad accounts is quite different from normal accounts.

%\begin{table}[h!]
%\begin{center}
%\begin{tabular}{|c|c|c|}
%\hline
 %Model & PR-AUC Median $\pm$ MAD \\ 
 %\hline
 %TIES-CNN & 0.8335  $\pm$ 0.0029 \\  
 %\hline
 %TIES-RNN & 0.8368 $\pm$ 0.0023 \\
 %\hline
 % TIES-Deepset & 0.7905 $\pm$ 0.0040 \\
 %\hline
 %Prod & 0.9040 $\pm$ 0.0022 \\
 %\hline
 % Prod+TIES-CNN & 0.9128 $\pm$ 0.0012 \\
 %\hline
  %Prod+TIES-RNN & 0.9111 $\pm$ 0.0021 \\
 %\hline
 %Prod+TIES-Deepset & 0.9100 $\pm$ 0.0021 \\
 %\hline
%\end{tabular}
%\end{center}
%\caption{Precision-Recall area under the curve median and median absolute deviation on the test dataset for \textbf{ads payment risks}. Combining TIES with prod improves the performance.}
%\label{table:payment_results_prauc}
%\end{table}
\begin{table}[h!]
\begin{center}
\begin{tabular}{|c|c|c|}
\hline
 Model & PR-AUC Median Gap $\pm$ MAD \\ 
 \hline
 TIES-CNN & -0.0705  $\pm$ 0.0051 \\  
 \hline
 TIES-RNN & -0.0672 $\pm$ 0.0045 \\
 \hline
  TIES-Deepset & -0.1135 $\pm$ 0.0062 \\
 \hline
  Baseline+TIES-CNN & 0.0088 $\pm$ 0.0034 \\
 \hline
  Baseline+TIES-RNN & 0.0071 $\pm$ 0.0043 \\
 \hline
 Baseline+TIES-Deepset & 0.0060 $\pm$ 0.0043 \\
 \hline
\end{tabular}
\end{center}
\caption{Median Precision-Recall area under the curve gap with respect to the baseline model as well as the median absolute deviation on the test dataset for \textbf{ads payment risks}. Combining TIES with the baseline improves the performance.}
\label{table:payment_results_prauc}
\end{table}

TIES models do a decent job in identifying bad accounts although they are not as predictive as the baseline model. Moreover, similar to the previous two experiments, RNN and CNN perform roughly the same and both outperform Deepset by a wide margin. Finally, combining TIES with the baseline provides about 1$\%$ gain in precision-recall area under the curve which is statistically significant. It is worth noting that even a small improvement in recall for the same precision would translate to a large monetary value in savings.
\section{Conclusion}~\label{sec:discussions}

In this paper, we  provided an overview of the temporal interaction embeddings (TIES) framework and demonstrated its effectiveness in fighting abuse at Facebook. Social media entities such as accounts, posts, Pages, and Groups interact with each other overtime. The type of interactions generated by bad entities such as fake accounts and hateful posts are different from normal entities. TIES, a supervised deeplearning  framework, embeds these interactions. The embedding vectors in turn can be used to identify bad entities (or improve existing classifiers). The TIES framework is quite general and can be used by various forms of account, post, Page, or Group integrity applications. Moving forward, we plan to continue exploring other applications of this methodology within Facebook. Moreover, we can add additional features such as model interpretability, hyperparameter optimization, unsupervised learning, etc. to the framework in order to create a more complete tool.

%ACKNOWLEDGEMENTS are optional
\section{Acknowledgements}
Authors would like to thank He Li, Bolun Wang, Yingyezhe Jin, Despoina Magka, Austin Reiter, Hamed Firooz, Shaili Jain for numerous helpful conversations, data preparation, and experiment setup.

\bibliographystyle{abbrv}
\bibliography{citations}  % sigproc.bib is the name of the Bibliography in this case
% ACM needs 'a single self-contained file'!
% SIGKDD: balancing columns messes up the footers: Sunita Sarawagi, Jan 2000.
% \balancecolumns

% \appendix
% \section{If needed}

% That's all folks!
\end{document}